\documentclass[10pt,twocolumn,letterpaper]{article}

\usepackage[pagenumbers]{cvpr} 
\usepackage{algorithm}
\usepackage{algorithmic}
\usepackage{indentfirst}
\usepackage{multirow}
\usepackage{graphicx}
\usepackage{xspace}
\usepackage{url}
\usepackage[marginal]{footmisc}
\usepackage[dvipsnames]{color}
\usepackage[dvipsnames]{xcolor}

\definecolor{cvprblue}{rgb}{0.21,0.49,0.74}
\usepackage[pagebackref,breaklinks,colorlinks,citecolor=cvprblue]{hyperref}

\def\paperID{9026} 
\def\confName{CVPR}
\def\confYear{2024}

\title{AtomoVideo: High Fidelity Image-to-Video Generation}

\author{
    Litong Gong$^{*}$, ~~%
    Yiran Zhu$^{*}$, ~~%
    Weijie Li$^{*}$, ~~%
    Xiaoyang Kang$^{*}$, ~~%
    Biao Wang, ~~%
    Tiezheng Ge, ~~%
    Bo Zheng ~~%
    \\
Alimama Tech, Alibaba Group\\
Beijing, China\\
{\tt\small \{gonglitong.glt, yizhu.zyr, weijie.lwj0, kangxiaoyang.kxy, }\\
{\tt\small eric.wb, tiezheng.gtz, bozheng\}@alibaba-inc.com}
}

\begin{document}
\twocolumn[{%
\renewcommand\twocolumn[1][]{#1}%
\maketitle
\begin{center}
    \centering
    \includegraphics[width=\linewidth,scale=1.0]{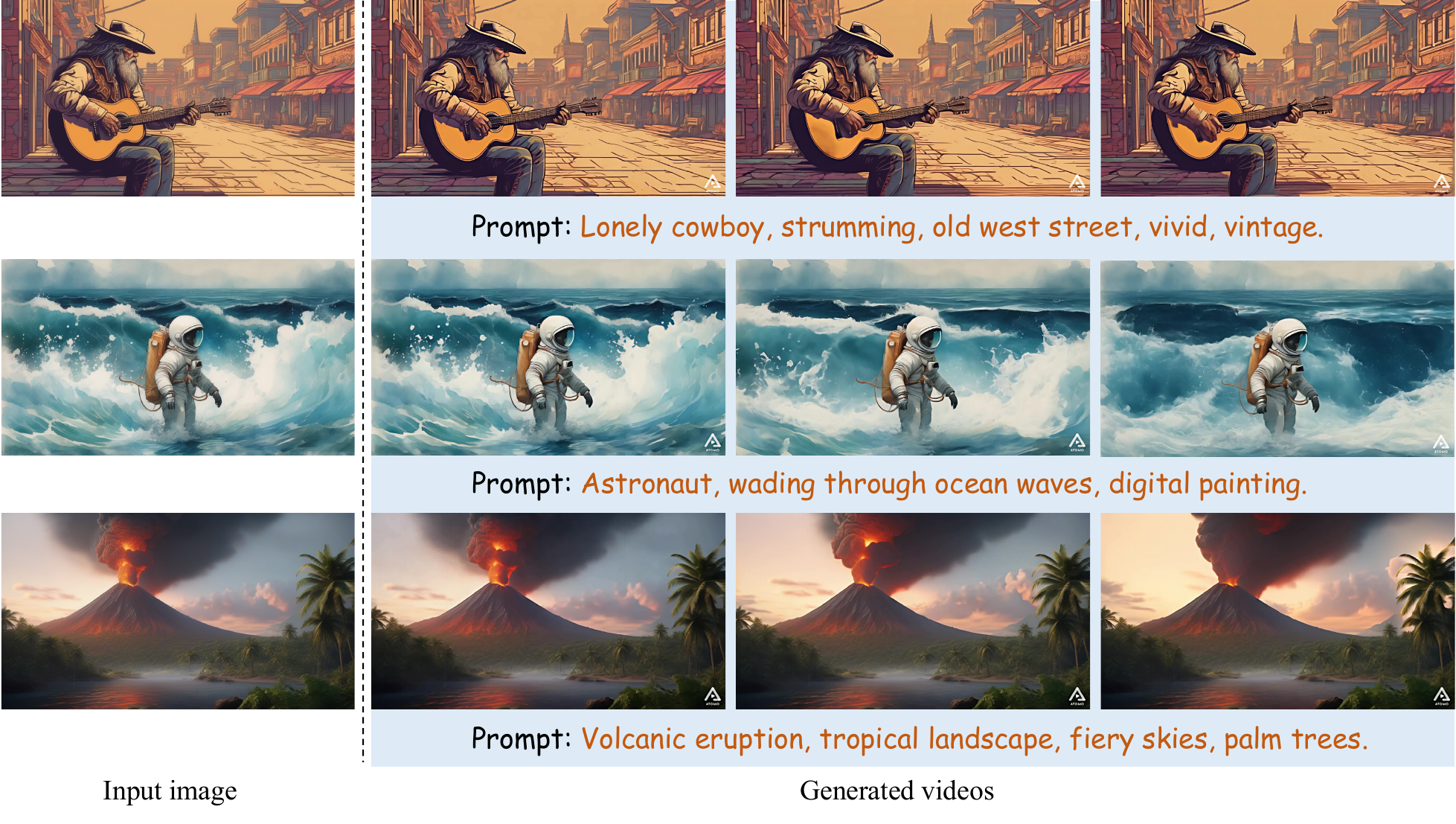}
    \captionof{figure}{Given a reference image and prompt, AtomoVideo can generates vivid videos while maintaining high fidelity detail with the given image.}
\end{center}%
}]
\footnote{$^{*}$These authors contributed equally to this work.}

\begin{abstract}
Recently, video generation has achieved significant rapid development based on superior text-to-image generation techniques. In this work, we propose a high fidelity framework for image-to-video generation, named \textbf{AtomoVideo}. Based on multi-granularity image injection, we achieve higher fidelity of the generated video to the given image. In addition, thanks to high quality datasets and training strategies, we achieve greater motion intensity while maintaining superior temporal consistency and stability. Our architecture extends flexibly to the video frame prediction task, enabling long sequence prediction through iterative generation. Furthermore, due to the design of adapter training, our approach can be well combined with existing personalised models and controllable modules. By quantitatively and qualitatively evaluation, AtomoVideo achieves superior results compared to popular methods, more examples can be found on our project website: \href{https://atomo-video.github.io/}{https://atomo-video.github.io/}.

\end{abstract}

\section{Introduction}
Recently, video generation based on diffusion models\cite{singer2022make, wang2023modelscope, blattmann2023align, guo2023animatediff, blattmann2023stable, dai2023emu}, have shown a growing interest and remarkable progress with impressive performance. In this paper, we introduce AtomoVideo, a novel framework for high-fidelity image-to-video(I2V) generation. AtomoVideo can generate high-fidelity videos from input image, achieving superior motion intensity and consistency compared to existing works. In combination with the advanced text-to-image(T2I) model\cite{rombach2022ldm, ramesh2022dalle2, saharia2022imagen, podell2023sdxl}, AtomoVideo also can achieve text-to-video(T2V) generation. In addition, our approach can be flexibly combined with personalised T2I models and controlled generative models\cite{zhang2023adding, mou2023t2i} for more customised and controllable generation, and we hope that AtomoVideo will contribute to the development of the video generation community.

Image-to-video generation is different from text-to-video generation because it requires to ensure as much as possible the style, content, and more fine-grained details of the given image, which greatly increases the challenge of the image-to-video generation task. Recently, an increasing number of researchers\cite{zhang2023i2vgen, blattmann2023stable, girdhar2023emu, zhang2023pia, chen2023livephoto, guo2023i2v} have focused on the area of image-to-video generation. In order to improve the consistency with the given image, some methods\cite{zhang2023i2vgen, blattmann2023stable, guo2023i2v} encode the image as high-level image prompts to inject into the model with cross-attention, such methods are difficult to achieve consistency of fine-grained details due to the utilisation of only higher-order semantics. In addition to this, a simpler idea is the concatenation of additional channels at the input, which although inputs more fine-grained low-level information, is harder to converge and generates poorer stability of the video. Therefore, a increasing number of works\cite{blattmann2023stable, chen2023livephoto} use both of the above methods for image information injection.However, some of these methods\cite{guo2023i2v, chen2023livephoto, dai2023fine} use a noisy prior instead of starting with pure Gaussian noise during inference, in order to compensate for the artifacts of model instability. Since the noise prior contains information of the given image, such as the inversion of the reference latent, the fidelity of the fine-grained details can be significantly enhanced. However, such methods significantly reduce the motion intensity, due to the fact that each frame contains exactly the same given image prior in the noise, making the initial noise random component decrease, which results in a reduction of the motion intensity.

In this work, to address the challenges presented above, our work presents an image-to-video generation model that achieves high fidelity and coherent motion without relying on noise priors. Specifically, we concatenate the given image at the input, while also injecting high-level semantic cues through cross-attention to improve the consistency of the video generation with the given image. During training, we employ zero terminal Signal-to-Noise Ratio\cite{lin2024common, girdhar2023emu} and v-prediction strategies\cite{salimans2022progressive}, which we analyse can significantly improve the stability of generation without a noisy prior. Moreover, our framework can be easily adapted to the video frame prediction task by predicting the following video frames, given the preceding frames, and through iterative generation, which enables the generation of long videos. Finally, we maintain a fixed T2I model during training, only adjusting the added temporal layer and input layer parameters, so it can be combined with the community's personalised T2I model and the controllable models for more flexible video generation.


\begin{figure*}[!ht]
\centering
\includegraphics[width=\linewidth,scale=1.0]{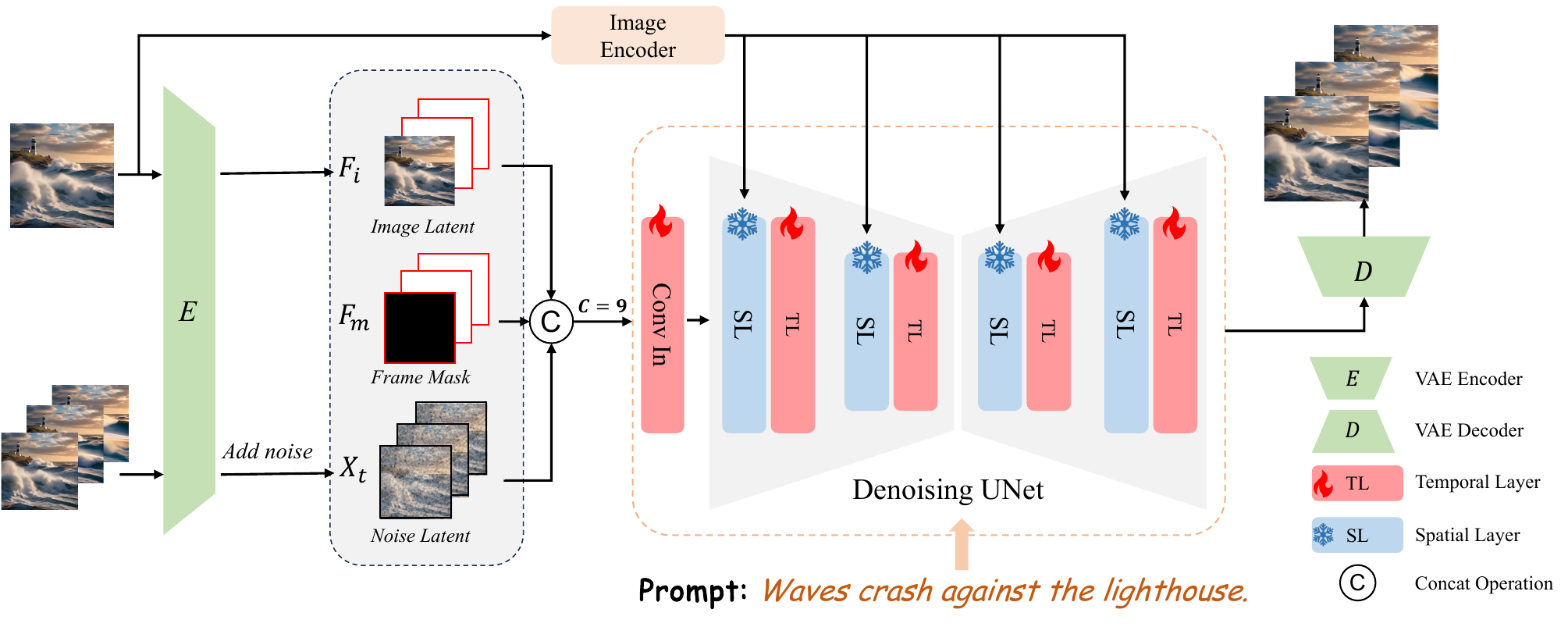}
\caption{The framework of our image-to-video method. During training, only the temporal and input layers are trained, and during testing, the noise latent is a sampled from Gaussian distribution without any reference image prior.}
\label{fig:framework}
\end{figure*}

\section{Related Work}
\textbf{Diffusion Models.} Due to the outstanding generative capabilities and controllability, Diffusion Probabilistic Model (DPM) \cite{ho2020dpm} and its variants have recently ascended to a dominant status within the field of generative modeling. 
Diffusion models \cite{ho2020dpm, sohl2015deep, dhariwal2021diffusion} accomplish the iterative refinement process by learning to progressively denoise samples from the normal distribution, while subsequent works \cite{rombach2022ldm, song2020score} reduce the computational burden by further leveraging learned representations in the latent space .
For text-to-image generation models \cite{rombach2022ldm, ramesh2022dalle2, saharia2022imagen, podell2023sdxl, dai2023emu}, it is common to use a language model such as CLIP \cite{radford2021learning} and T5 \cite{raffel2020exploring} as a text encoder and introduce it by means of cross-attention \cite{vaswani2017attention} to improve the alignment of text and images.
Beyond natural language inputs, the use of additional image conditions to guide the layout of the generated images \cite{zhang2023adding, mou2023t2i, huang2023composer} also becomes an active area of research.

\textbf{Text-to-Video Synthesis with Diffusion Models.}
As diffusion models have prospered in image generation tasks, the use of diffusion models for video generation has received increasing attention. 
Early attempts \cite{singer2022make, wang2023modelscope, blattmann2023align} focused on generating videos from text by adding a time dimension to text-to-image models, allowing them to capture temporal information.
AnimateDiff \cite{guo2023animatediff} learns a plug-and-play motion module from large-scale video data by keeping the original weights of the text-to-image model fixed.
To enhance the usability of the results, some works have improved the quality of generated videos by leveraging the diffusion noise prior \cite{ge2023preserve} or cascading models \cite{ho2022imagen, wang2023lavie}.
Additionally, controllable video generation is also an important area. Some work have incorporated additional control signals like depth maps \cite{esser2023structure}, human poses \cite{ma2023follow}, or a combination of multiple conditions \cite{zhang2023controlvideo, wang2023videocomposer} to create videos that more accurately meet user needs.

\textbf{Image-to-Video Synthesis with Diffusion Models.}
Recently, image-to-video generation has been emerging as an active area of research. This field not only focuses on the overall quality of generated content but also pays attention to the fidelity of the input image and the plausibility of the motion effects.
I2VGen-XL \cite{zhang2023i2vgen} achieves high-resolution image-to-video generation by decoupling the tasks of semantic scene creation and detail enhancement through two cascaded models.
Stable Video Diffusion  \cite{blattmann2023stable} leverages text-to-video pre-training on a carefully curated dataset to enable the model to learn strong motion priors, which are further applied to downstream tasks like image-to-video and multi-view synthesis. 
Emu Video \cite{girdhar2023emu} learns to directly generate high-quality outputs by adjusting the noise scheduling and employing multi-stage training.
Some works \cite{zhang2023pia, chen2023livephoto} incorporates additional input channels to bolster control over the overall intensity of the video's motion effects.
In our work, we focus not only on the high-fidelity consistency of the given image but also achieve high-quality motion effects. More importantly, as we have fixed the weights of the spatial layers, our work can seamlessly integrate with existing plugins such as ControlNet \cite{zhang2023adding}, LoRAs \cite{hu2022lora}, and stylized base models.
\section{Method}

\subsection{Overall Pipeline}
Our overall process is shown in Fig.\ref{fig:framework}, We use the pre-trained T2I model, newly added 1D temporal convolution and temporal attention modules after every spatial convolution and attention layer, with fixed T2I model parameters and only training the added temporal layer. Meanwhile, in order to inject the image information, we modify the input channel to 9 channels, add the image condition latent and binary mask. Since the input concatenate image information is only encoded by VAE, it represents low-level information, which contributes to the enhancement of fidelity of the video with respect to the given image. Meanwhile, we also inject high-level image semantic in the form of cross-attention to achieve more semantic image controllability.

\subsection{Image Information Injection}
Currently, with the rapid development of diffusion models, text-to-image generation has enabled the generation of highly aesthetic creative images. Therefore, achieving high-quality image-to-video video generation based on a given image is a popular research topic. In which, consistency preservation with a given image and video motion coherence in I2V tasks are usually trade-offs. In our approach, images are injected at two separate positions. As shown in Fig.\ref{fig:framework}, we encode the image through VAE encoder to obtain the low-level representation, formulated as $F_{i}$, and the corresponding input frame mask $F_{m}$, $F_{i}$ and $F_{m}$ are concatenated with the Gaussian noise $X_{t}$ in the channel dimension, described by the formula:

$$
X_{t}^{'}=Concat(X_{t}, F_{m}, F_{i}),
$$
Where $X^{'}_{t}$ is the final input to the UNet with channel dimension $C=9$. The image condition $F_{i}$ contains such information that can recover fine-grained image details, which is extremely important for the fidelity of the generated video to the given image. 

In addition, we simultaneously encode the input image with a CLIP image encoder\cite{radford2021learning-CLIP} to yield the high-level semantic representation in patch granularity, which is followed by a linear projection layer for dimension transformation and injected through the added cross-attention layer. In the detailed implementation, we used IP-Adapter\cite{14_IP-Adapter_ye2023ip-adapter} based on SD1.5\cite{rombach2022ldm} pre-trained model weights for training.

\begin{figure}[tbp]
\centering
\includegraphics[width=\linewidth,scale=1.0]{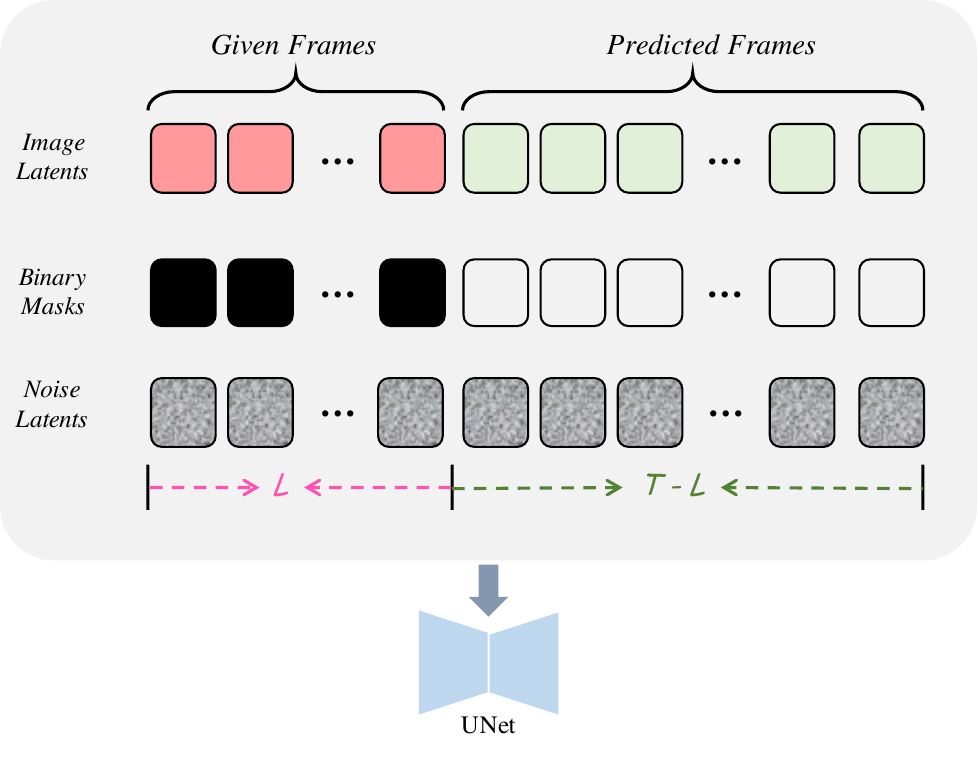}
\caption{Illustration of video prediction. Given a length $L$ sequence of video frames, predicting the subsequent frames of $T-L$ is performed by making adaptation only at the input layer, with no additional adjustment of the model. And $T$ denotes the maximum sequence of frames supported by the model.}
\label{fig:video-prediction}
\end{figure}

\subsection{Video Frames Prediction}
Long video generation is a significant challenge in video diffusion models due to the constraints of GPU memory. We extend our approach to the task of video frame prediction by implementing long video generation in an iterative manner by predicting subsequent frames given the preceding frames. Specifically, the input image conditions, image latents $F_{i}$ and frame mask $F_{m}$ in Fig.\ref{fig:framework}, can be flexibly replaced with any several frames from a given video, as illustrated in Fig.\ref{fig:video-prediction}. Typically, for video frame prediction, we input the first $L=8$ frames to the model and predict the subsequent $T-L=16$ frames. Apart from that, the model structure does not require any other changes. We use the well-trained I2V model as an initialisation, and train it with only a little number of videos to converge quickly and achieve relatively stable long video generation.

\subsection{Training and Inference}
We employ Stable Diffusion 1.5 as our foundational Text-to-Image (T2I) model and initialize the temporal attention layers with AnimateDiff. We use our 15M internal dataset for training, where each video is about 10-30 seconds in length and the textual description of the video is also fed into the model. In addition, we employ zero terminal Signal-to-Noise Ratio (SNR)\cite{lin2024common} and v-prediction\cite{salimans2022progressive} when training, which in our practice proved that they are effective on the stability of video generation. The input size of our model is $512\times512$ and contains $24$ frames.

During inference, We perform Classifier-Free Guidance\cite{ho2022classifier} with both image and text conditional injection. Our empirical validation indicates that image conditions combined with text prompts significantly increase the stability of the generated output.


\begin{table*}[htbp]
\centering
\scalebox{1.0}{%
\begin{tabular}{cccccc}
\hline
\multirow{2}{*}{Methods} &
  \begin{tabular}[c]{@{}c@{}}Image\\ Consistency\end{tabular} &
  \begin{tabular}[c]{@{}c@{}}Temporal \\ Consistency\end{tabular} &
  \begin{tabular}[c]{@{}c@{}}Video-Text\\ Alignment\end{tabular} &
  \begin{tabular}[c]{@{}c@{}}Motion\\ Effects\end{tabular} &
  \begin{tabular}[c]{@{}c@{}}Video\\ Quality\end{tabular} \\ \cline{2-6} 
              & SSIM$\uparrow$           & ICS$\uparrow$             & CLIP Score$\uparrow$     & RAFT$\uparrow$           & DOVER$\uparrow$          \\ \hline
VideoCrafter\cite{chen2023videocrafter1}  & 0.417          & 0.9906          & 0.259          & 0.384          & 0.601          \\
I2VGEN-XL\cite{zhang2023i2vgen}     & 0.417          & 0.9795          & 0.248          & 1.271          & 0.552          \\
SVD\cite{blattmann2023stable}            & 0.615          & 0.9830          & 0.273          & 2.778          & 0.726          \\
Pika\cite{pika}          & 0.739          & \textbf{0.9974} & 0.274          & 0.192          & 0.747          \\
Gen-2\cite{gen2}          & \textbf{0.835} & 0.9972          & 0.274          & 0.497          & \textbf{0.824} \\ \hline
\textbf{Ours} & 0.759          & 0.9938          & \textbf{0.279} & \textbf{3.124} & 0.804          \\ \hline
\end{tabular}
}
\caption{Quantitative comparison of AtomoVideo with other methods.}
\label{tab:my-table}
\end{table*}

\begin{figure*}[!ht]
\centering
\includegraphics[width=\linewidth,scale=1.0]{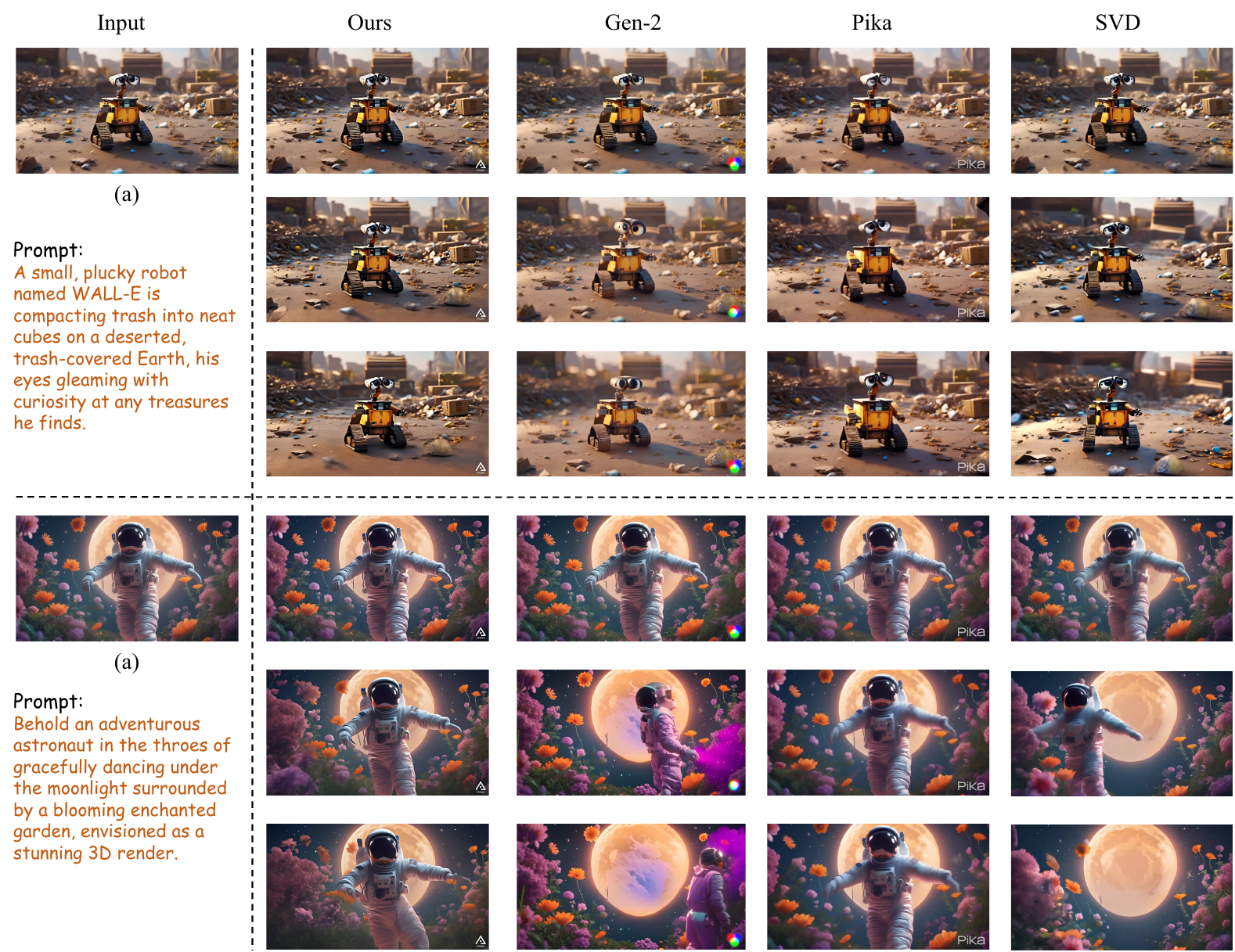}
\caption{Samples comparison with other methods. We compare the SVD\cite{blattmann2023stable}, Pika\cite{pika} and Gen-2\cite{gen2}, where AtomoVideo maintains better stability and greater motion intensity.
}
\label{fig:compare}
\end{figure*}

\section{Experiments}
\subsection{Quantitative Comparisons}

\begin{figure*}[]
\centering
\includegraphics[width=\linewidth,scale=1.0]{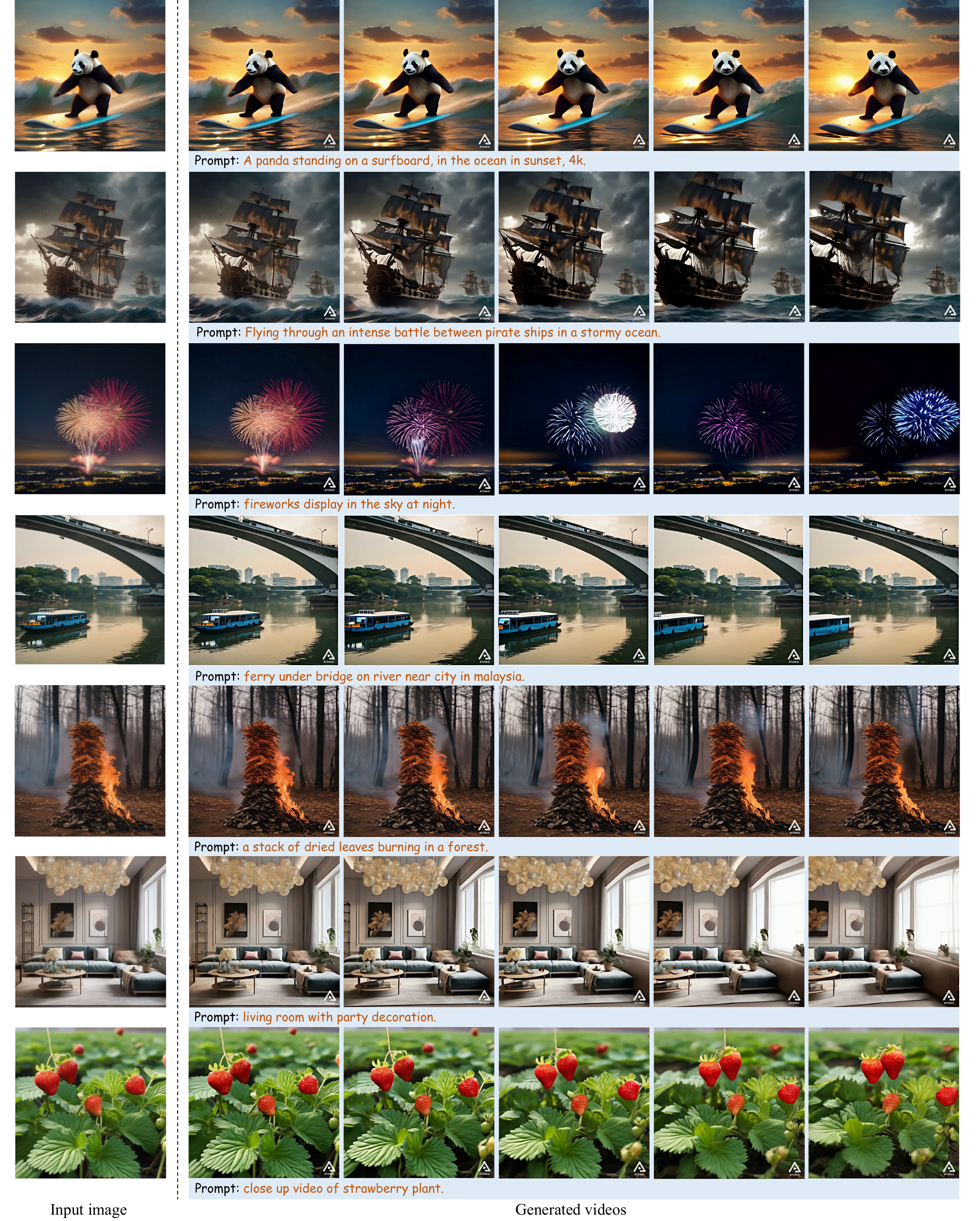}
\caption{More samples with $512\times512$ size.}
\label{fig:case512}
\end{figure*}

\begin{figure*}[]
\centering
\includegraphics[width=\linewidth,scale=1.0]{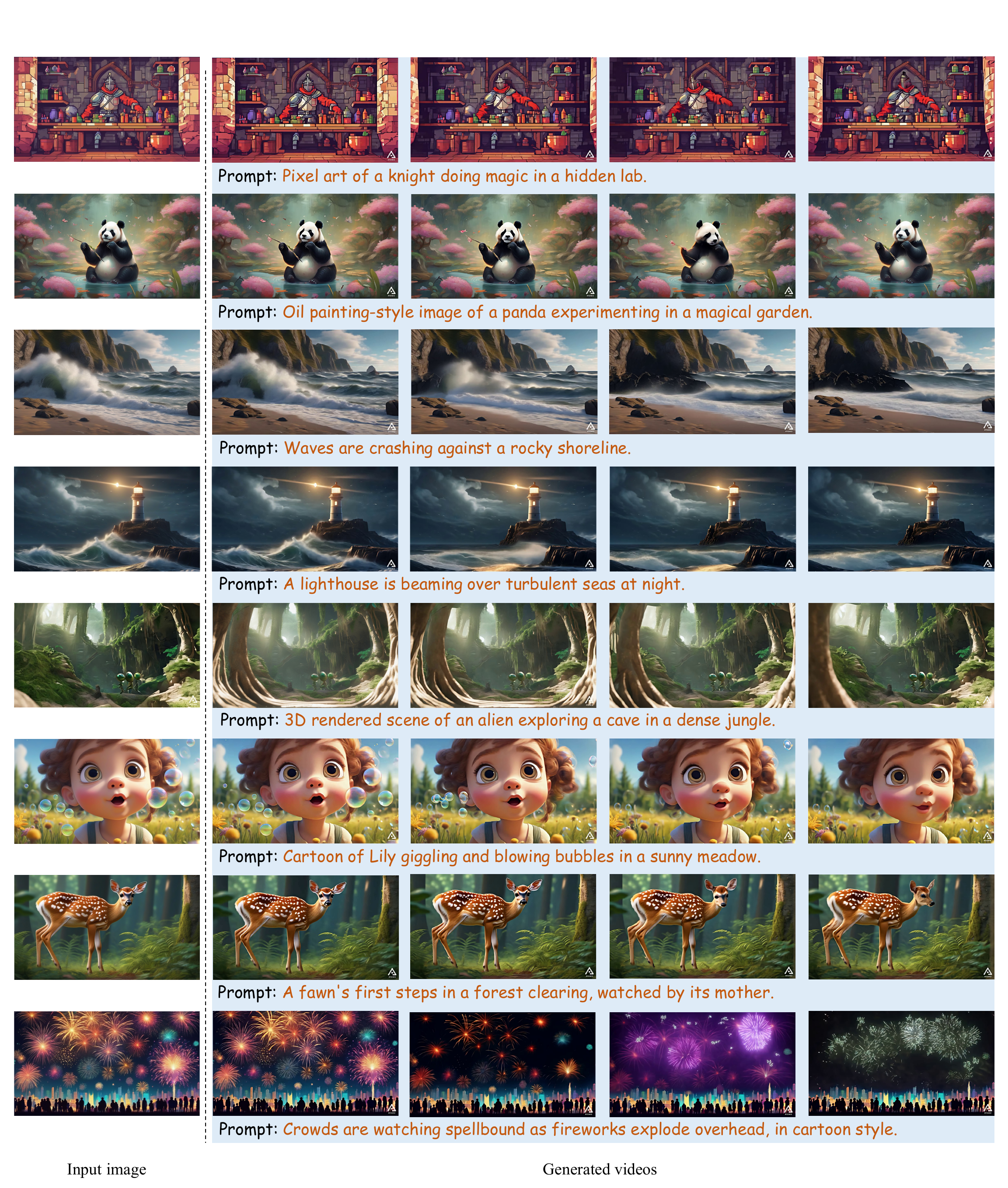}
\caption{More samples with $1280\times720$ size.}
\label{fig:case1280}
\end{figure*}

\textbf{Evaluation Setting.} We follow the AIGCBench\cite{fan2024aigcbench} setting for evaluation, which provides more comprehensive evaluation criterions in the I2V task. We compare recent excellent methods in the I2V domain such as VideoCraft\cite{chen2023videocrafter1}, I2VGEN-XL\cite{zhang2023i2vgen}, SVD\cite{blattmann2023stable}, and also commercial methods such as Pika\cite{pika} and Gen-2\cite{gen2}. We calculate metrics on multiple dimensions commonly used in the field, including 1).Image Consistency. We calculate Structural Similarity Index Measure(SSIM)\cite{wang2004ssim} between the first frame of the generated video and the reference image to evaluate the generation fidelity with the given image. 2).Temporal Consistency. We compute the image CLIP\cite{radford2021learning-CLIP} score(ICS) between adjacent frames of the generated video to measure temporal coherence and consistency. 3). Video-Text Alignment. We use the CLIP\cite{radford2021learning-CLIP} score of the video frames to the prompt to measure the degree of video-text alignments. 4). Motion Intensity. To avoid over-optimising the image fidelity in preference to generating static videos, we use RAFT calculate the flow score between adjacent frames of the generated video to represent the magnitude of the motion intensity. 5). Video Quality. We utilize disentangled objective video quality evaluator(DOVER)\cite{wu2023DOVER} to evaluate the video quality.

\textbf{Quantitative Results.} The quantitative evaluation results are shown in Table~\ref{tab:my-table}, comparing with other excellent open source methods, including VideoCrafter\cite{chen2023videocrafter1}, I2VGEN-XL\cite{zhang2023i2vgen} and SVD\cite{blattmann2023stable}, we achieve the best scores in all evaluation dimensions, especially in image consistency. Besides, comparing with the commercial methods, we also show advantages in several dimensions, especially the motion intensity score. AtomoVideo shows greater motion intensity(RAFT) with competitive temporal consistency compared to Pika\cite{pika} and Gen-2\cite{gen2}, while they tend to generate static videos. Further, it is worth noting that we are slightly lower than commercial methods in image consistency and video quality, we analyse two reasons for this, one is the influence of the resolution of the generated video, and the other is that they may employ a better base model, whereas we utilize SD-1.5 and fix the parameters, and we believe that we can obtain a superior video by employing more advanced base models.

\begin{figure}[tbp]
\centering
\includegraphics[width=\linewidth,scale=1.0]{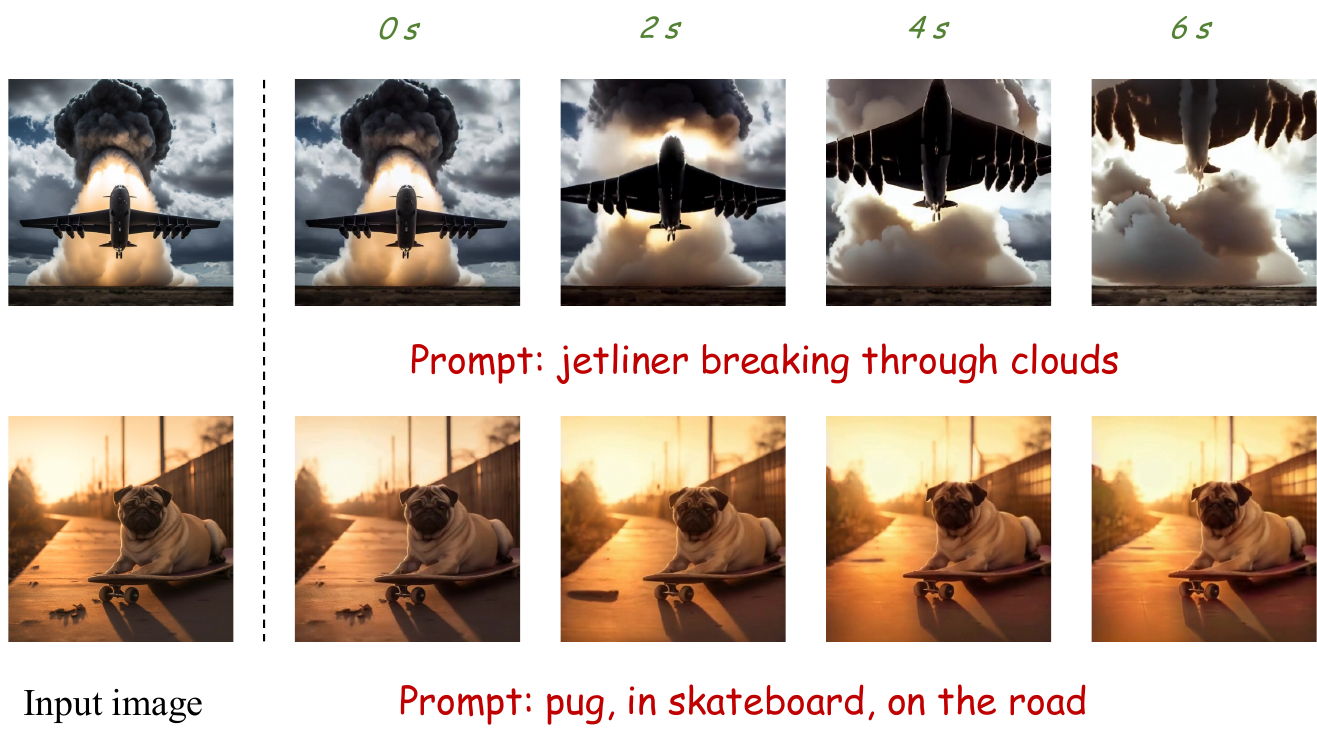}
\caption{Samples of long video generation. The left is the input image and the right is the generated video of 7s length.}
\label{fig:long-video-case}
\end{figure}

\subsection{Qualitative Samples}
In this section, we show some qualitative samples in Fig.\ref{fig:compare}. We compare our method with SVD\cite{blattmann2023stable}, the commercial methods Pika\cite{pika} and Gen-2\cite{gen2}, which all achieve relatively high fidelity with the given image. However, in our experiments, pika is more preferred to generate static videos, while the results generated by Gen-2\cite{gen2} and SVD\cite{blattmann2023stable} are susceptible to artifacts when the subject undergoes a large motion change. Overall, compared to other methods, we achieve more coherent and stable temporal consistency when generating videos with greater motion intensity. We train our model on $512\times512$ size and more examples are shown in Fig~\ref{fig:case512}, but we find also great generalisations on other resolutions, e.g. some samples generated on $1280\times720$ size are shown in the Fig~\ref{fig:case1280}.

Besides, as shown in Fig.\ref{fig:long-video-case}, demonstrating the results of video frame prediction, we achieve longer video generation by iterative video frame prediction.

\begin{figure}[tbp]
\centering
\includegraphics[width=\linewidth,scale=1.0]{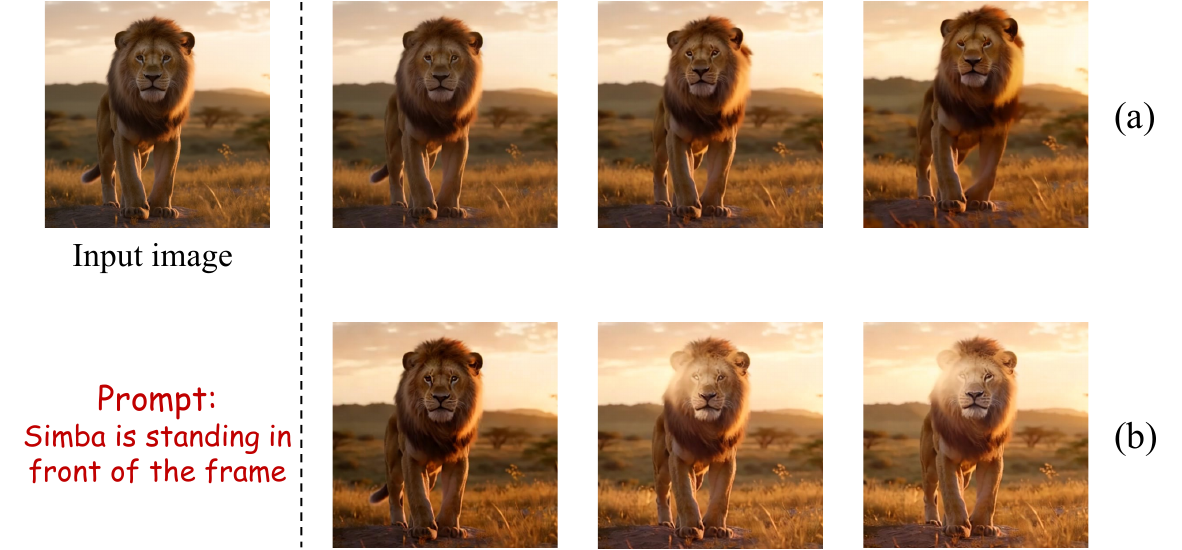}
\caption{Comparison using SD-1.5 and \href{https://civitai.com/models/25694/epicrealism}{epiCRealism} models. (a). Results generated using the SD-1.5 model, consistent with training. (b). Results generated using the epiCRealism model, with image-to-video generation injected with more light elements.}
\label{fig:epic}
\end{figure}

\subsection{Personelized Video Generation}
Since our method freezes the parameters of the base 2D UNet and trains only the added parameters, it can be combined with the popular personalised models in the community. As shown in Figure\ref{fig:epic}, we show the results of combining our model with \href{https://civitai.com/models/25694/epicrealism}{epiCRealism}, a T2I model that is excellent for light and shadow generation, and utilizing it for I2V generation prefers to generate videos with light elements. In this work, since we emphasise more on the fidelity of the generated video with respect to the given image, it is not easy to work in combination with many stylistic models such as cartoon style.

\section{Conclusion}
In this work, we present AtomoVideo, a high-fidelity image-to-video generation framework. Our method greatly exploits the generative capabilities of the T2I model and is trained only on the parameters of the added temporal and input layers. Qualitative and quantitative evaluations indicate that our method achieves excellent performance, maintaining superior temporal consistency and stability in the case of generating video with greater motion intensity. In the future, we will work towards more controllable image-to-video generation, as well as expanding to more powerful T2I base models.
{
    \small\bibliographystyle{ieeenat_fullname}
    \bibliography{main}
}


\end{document}